\documentclass{article}

\PassOptionsToPackage{numbers, compress}{natbib}
\usepackage[final]{neurips_2024}

\usepackage[utf8]{inputenc} 
\usepackage[T1]{fontenc}    
\usepackage{hyperref}       
\usepackage{url}            
\usepackage{booktabs}       
\usepackage{amsfonts}       
\usepackage{nicefrac}       
\usepackage{microtype}      
\usepackage{xcolor}         
\usepackage{amsmath,amssymb}
\usepackage{graphicx}
\usepackage{wrapfig}

\newcommand{\eg}{\textit{e.g.}}
\newcommand{\ie}{\textit{i.e.}}

\newcommand{\R}{\mathbb{R}}

\newcommand{\dd}{d_\mathrm{data}}

\newcommand{\dr}{d_\mathrm{repr}}

\newcommand{\drmax}{\dr^{{\mathrm{max}}}}
\newcommand{\argmax}{\mathrm{argmax}}

\title{Pre-processing and Compression: Understanding Hidden Representation Refinement Across Imaging Domains via Intrinsic Dimension}

\author{Nicholas Konz$^{1}$, Maciej A. Mazurowski$^{1,2,3,4}$ \\
$^{1}$ Department of Electrical and Computer Engineering, $^{2}$ Department of Radiology, \\
$^{3}$ Department of Computer Science, $^{4}$ Department of Biostatistics \& Bioinformatics \\
Duke University, NC, USA \\
\texttt{\{nicholas.konz, maciej.mazurowski\}@duke.edu} \\
}

\begin{document}

\maketitle

\begin{abstract}
In recent years, there has been interest in how geometric properties such as \textit{intrinsic dimension} (ID) of a neural network's hidden representations change through its layers, and how such properties are predictive of important model behavior such as generalization ability.
However, evidence has begun to emerge that such behavior can change significantly depending on the domain of the network's training data, such as natural versus medical images. Here, we further this inquiry by exploring how the ID of a network's learned representations changes through its layers, in essence, characterizing how the network successively refines the information content of input data to be used for predictions. 
Analyzing eleven natural and medical image datasets across six network architectures, we find that how ID changes through the network differs noticeably between natural and medical image models. Specifically, medical image models peak in representation ID earlier in the network, implying a difference in the image features and their abstractness that are typically used for downstream tasks in these domains. 
Additionally, we discover a strong correlation of this peak representation ID with the ID of the data in its input space, implying that the intrinsic information content of a model's learned representations is guided by that of the data it was trained on. Overall, our findings emphasize notable discrepancies in network behavior between natural and non-natural imaging domains regarding hidden representation information content, and provide further insights into how a network's learned features are shaped by its training data.
\end{abstract}

\section*{Introduction}

In science, it is common to use relatively simple models to approximate the behavior of complex systems. The parameters of such models for real systems can often be estimated using observed data, which can then be used to extrapolate to future behavior. This approach has been increasingly applied to deep learning, such as where relatively simple measurable geometrical properties (intrinsic dimension, curvature, etc.) of a network's manifold of learned representations \citep{ansuini2019intrinsic} or training data \citep{pope2021intrinsic} can predict important behavior such as generalization, adversarial robustness, or prompt perplexity \citep{kvinge2023exploring}. However, it has been recently found that such behavior can vary significantly between data domains---in particular natural versus medical images---with many open questions remaining about what other behavior can change due to such domain shift \citep{konz2022intrinsic,konz2024effect,valeriani2024geometry}. Here, we further this inquiry by exploring how the \textit{intrinsic dimension} (ID) of a neural network's learned representations---which describes their intrinsic information content/minimum number of degrees of freedom---changes through its layers, and how this behavior changes depending on the dataset or data domain.

It has been found that for ImageNet \citep{deng2009imagenet} classification networks (both convolutional and transformer-based), the ID of hidden representations typically first steadily increases through successive network layers, then past a certain point, decreases until the output \citep{ansuini2019intrinsic,valeriani2024geometry}. It is hypothesized that this first stage of ``dimensionality expansion'' is the network \textit{pre-processing} the data to disentangle certain correlated, nuisance features (such as luminance or contrast) which are irrelevant to the final prediction, which may have parallels in human visual processing \citep{stringer2019high,tafazoli2017emergence}. Once this is completed and the ID has peaked, the second stage begins where the network successively \textit{compresses} the representations to lower- and lower-dimensional manifolds that are better suited for generalizing to unseen data \citep{ansuini2019intrinsic,ma2018dimensionality}. Importantly, prior results with ImageNet models have shown the ID of the final hidden layer representations to be closely correlated with generalization error \citep{ansuini2019intrinsic}, pointing it to being a useful property for describing network behavior. However, the generality of such findings in other datasets or data domains is unknown.

In this work, we first explore how layer-by-layer representation ID progression differs for models trained on datasets beyond ImageNet, including both three other natural image datasets and seven medical image datasets for a wide range of diagnostic tasks. We find that these curves have typically different shapes depending on if the model was trained on natural or medical images. In particular, the depth at which the ID peaks at some $\drmax$ is noticeably deeper into the model on average for natural image models, which we attribute to the task-relevant features of medical images typically requiring less abstract representations to capture, and so can be ``pre-processed'' for compression earlier in the network. Next, we explore the relationship between $\drmax$ and the intrinsic dimension of the raw data in its native space, $\dd$ \citep{pope2021intrinsic}, and discover a new result: there is a strong correlation between $\drmax$ and $\dd$ across all datasets and models, implying that the maximum intrinsic complexity of learned representations in a model is guided by the intrinsic complexity of the model's training data.

\section{Experimental Methods}


\paragraph{The manifold hypothesis and intrinsic dimension.}
Neural networks work by mapping high-dimensional input data (like images) to significantly lower-dimensional manifolds of learned representations \citep{gong2019intrinsic,ansuini2019intrinsic} which describe features that generalize to new data, made possible by learning the lower-dimensional manifold structure of such datasets in their native space \cite{fefferman2016manifoldhyp}. The intrinsic dimension of such a manifold can be computed via a relatively simple estimator that utilizes maximum likelihood estimation (MLE) \citep{levina2004maximum,mackay2005comments} given the high dimensional dataset (in our case, of representations) which we use here following prior work \citep{konz2024effect,pope2021intrinsic} via the implementation of \url{https://github.com/mazurowski-lab/intrinsic-properties}. We set the nearest-neighbor hyperparameter to $k=20$ following \cite{konz2024effect,konz2022intrinsic,pope2021intrinsic}. 

Consider some $C$-class image classification neural network $f:\R^n\rightarrow \R^C$ with $L$ layers $f_i$ indexed by $i=1,\ldots, L$, \ie, $f = f_L \circ\ldots \circ f_2 \circ f_1$, and $f_{\leq i}(x):= (f_i \circ\ldots \circ f_1)(x)$ denotes the $i^{th}$ hidden layer output given input data $x\in \R^n$. Given some \textit{unlabeled} input dataset $\mathcal{X}$, the intrinsic dimension $\dr^i$ of an $i^{th}$ layer's manifold of representations $\mathcal{H} := \{ f_{\leq i}(x): x \sim \mathcal{X}\}$ can be computed by applying one of the aforementioned estimators to $\mathcal{H}$. 
In this work, we will mainly analyze the way that representation ID changes from layer to layer, \ie, the sequence $\dr^1, \ldots, \dr^i,\ldots, \dr^{L}$.

\paragraph{Datasets.}
We experiment on binary classification datasets from both natural imaging and medical imaging. The former includes \textbf{ImageNet} \citep{deng2009imagenet}, \textbf{CIFAR10} \citep{krizhevsky2009cifar}, \textbf{SVHN} \citep{netzer2011svhn}, and \textbf{MNIST} \citep{deng2012mnist}, where the two predictive classes are randomly chosen for any given experiment (\ie, model training and ID estimation). The latter utilizes the same datasets and predictive tasks as \citep{konz2024effect}; including (1) brain MRI glioma detection (\textbf{BraTS}, \citep{menze2014multimodal}); (2) breast MRI cancer detection (\textbf{DBC}, \citep{saha2018machinedukedbc}); (3) prostate MRI cancer risk scoring (\textbf{Prostate MRI}, \citep{sonn2013prostate}); (4) brain CT hemorrhage detection (\textbf{RSNA-IH-CT}, \citep{flanders2020rsnaihct}); (5) chest X-ray pleural effusion detection (\textbf{CheXpert}, \citep{irvin2019chexpert}); (6) musculoskeletal X-ray abnormality detection (\textbf{MURA}, \citep{rajpurkar2017mura}); and (7) knee X-ray osteoarthritis detection (\textbf{OAI}, \citep{Tiulpin2018}).

We follow the same procedure for dataset creation as in \cite{konz2024effect}, creating randomly-sampled, even class-balanced training sets of size $N\in\{500, 750, 1000, 1250, 1500, 1750\}$ and test sets of size $750$; natural image model results are averaged over $5$ runs of different randomly selected class pairings. Representation ID estimates for a given model are made using its training set, following \cite{ansuini2019intrinsic}. Our results will be presented for $N=1750$ unless otherwise stated, but we provide all results for the other $N$ in Appendix \ref{app:additionalresults}. All images are resized to $224\times 224$ and normalized to $[0,1]$.

\paragraph{Models.}
We evaluate six common convolutional network architectures: VGG-13, -16, -19 \citep{VGGs} and ResNet-18, -34, and -50 \citep{he2016resnet}. Each network is trained via Adam \citep{adam} until maximally fitting to the given training set. We analyze the representations outputted by all pooling layers, convolutional blocks (or residual blocks for the ResNet models), and fully-connected layers.

\section{Experiments and Results}

\paragraph{Progression of hidden representation intrinsic dimension through network depth.}

\begin{figure}[!htbp]
\centering
\includegraphics[width=0.99\textwidth]{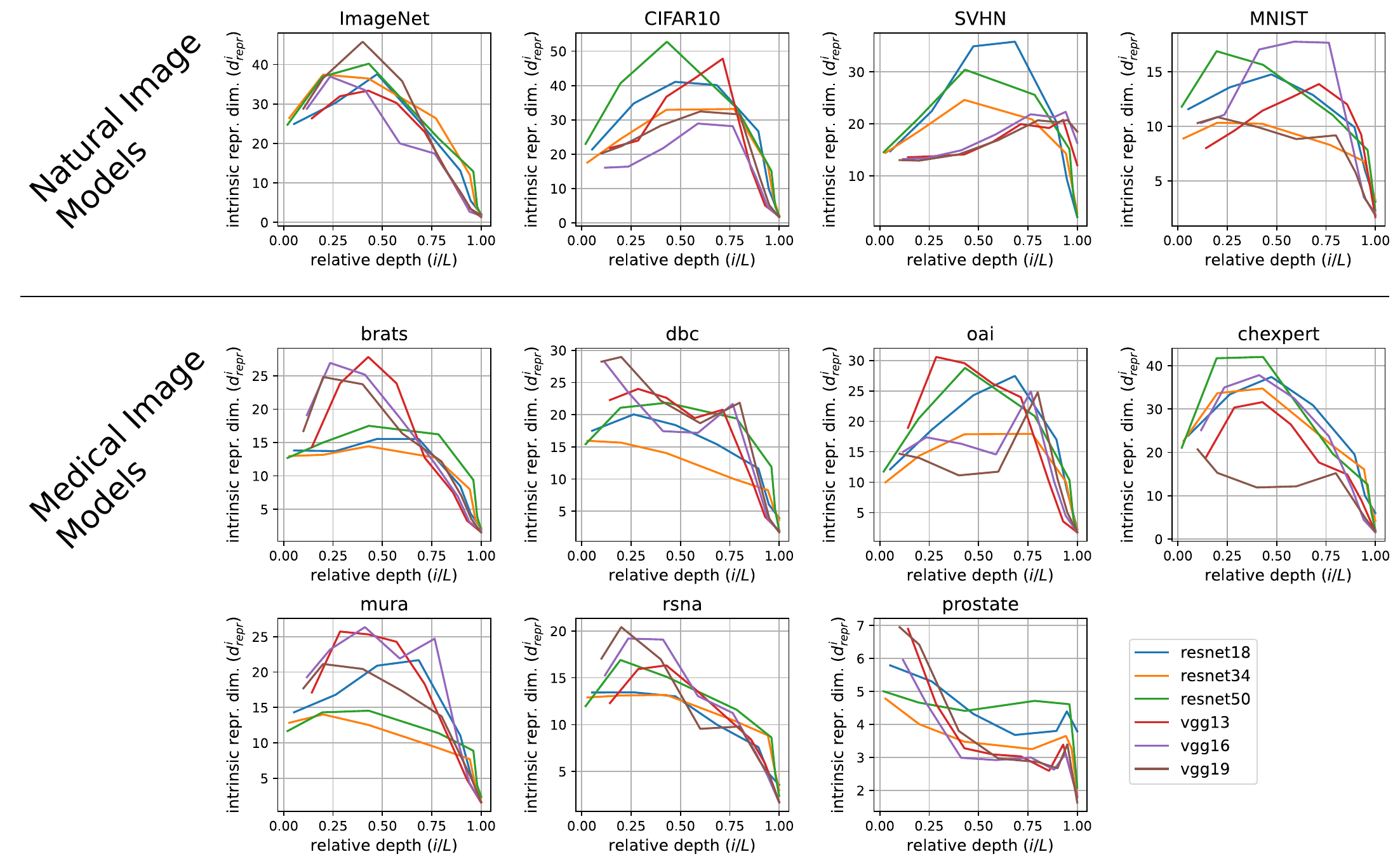}
\caption{Progression of representation intrinsic dimension $\dr^i$ with respect to relative network depth $i/L$, for all models trained on each dataset.}
\label{fig:humpbacks}
\end{figure}

We will first analyze how representation intrinsic dimension (ID) $\dr^i$ changes with respect to relative network depth $i/L$ for models trained on either natural or medical image datasets or, in other words, the way that intrinsic representation information content changes through the depth of the network. We will characterize these ID curves by the depth
\begin{equation}
    i^* := \argmax_{i} (\dr^i)
\end{equation}

at which the ID peaks at some $\drmax:= \dr^{i^*}$, transitioning from the aforementioned ``pre-processing'' stage to the ``compression'' stage of the model. This $\drmax$ is hypothesized to be the minimum number of degrees of freedom needed to represent the data such that task-irrelevant features are decorrelated with task-related features before the compression phase can begin. We show these curves for all datasets in Fig. \ref{fig:humpbacks}, along with peak ID values $\drmax$ and their depths $i^*/L$ for each dataset in Table \ref{tab:peaks}, averaged over all models.

\begin{table}[!htbp]
\centering
\scriptsize
\setlength{\tabcolsep}{4pt}
\begin{tabular}{l||cccc|c}
    & ImageNet & CIFAR-10 & SVHN & MNIST & \textbf{Avg.} \\
    \hline
    $\drmax$ & $39 \pm 4$ & $39\pm 9$ & $26\pm 6$ & $14\pm 3$ & $\mathbf{29\pm 10}$ \\
    $i^*/L$ & $0.40\pm 0.10$ & $0.65\pm 0.14$ & $0.77\pm 0.24$ & $0.47\pm 0.24$ & $\mathbf{0.57\pm 0.15}$ \\ \hline
\end{tabular}
\begin{tabular}{ccccccc|c}
    BraTS & DBC & OAI & CheXpert & MURA & RSNA-IH-CT & Prostate-MRI & \textbf{Avg.} \\
    \hline
    $21\pm 5$ & $23\pm 5$ & $26\pm 4$ & $34\pm 7$ & $21\pm5$ & $17\pm3$ & $6\pm1$ & $\mathbf{21\pm 8}$ \\
    $0.40\pm 0.10$ & $0.23\pm 0.14$ & $0.65\pm 0.22$ & $0.40\pm 0.16$ & $0.40\pm 0.16$ & $0.33\pm 0.10$ & $0.05\pm 0.00$ & $\mathbf{0.35\pm 0.17}$ \\
    \hline
\end{tabular}

\caption{Hidden representation peak intrinsic dimension $\drmax$ and the relative network depth $i^*/L$ at which it was attained for each dataset (see Fig. \ref{fig:humpbacks} for reference), averaged over all architectures. \textbf{Top table:} natural image datasets. \textbf{Bottom table:} medical image datasets. Confidence intervals provided as standard deviations over all model architectures.}
\label{tab:peaks}
\end{table}

We see that the typical shapes of these ID curves differ depending on whether the model was trained on a natural image or medical image dataset; in particular, the average $\drmax$ for natural image models is $0.57$, while for medical image models it is noticeably earlier in the networks at $0.35$. This implies that for medical image datasets, features needed for diagnostic tasks are typically less abstract/hierarchical than those for natural image tasks and so can be pre-processed by the network for prediction usage at earlier layers. We find the same for other training set sizes (Appendix \ref{app:additionalresults:humpbacks}).

\paragraph{The relationship between hidden representation intrinsic dimension and dataset intrinsic dimension.}

We hypothesize that the learned representation manifolds of a network may be shaped by the properties and complexity of the dataset manifold in its native (pixel) space. Prior work \citep{konz2024effect} studied the relationship between the intrinsic dimension of final hidden layer representations for classification networks, $\dr^{L-1}$, with the intrinsic dimension of the dataset itself, $\dd$, and theoretically and empirically showed that $\dr^{L-1} \lesssim \dd$. Intuitively, this describes that the network learns to convert the data to a final hidden representation constructed mainly of intrinsic features that are \textit{task-relevant}, which will be a subset of a larger number of intrinsic features that can be used to describe the data in its original space. Here we will explore this relationship between the intrinsic information content of input data and representations further.

\begin{wrapfigure}[22]{r}{0.45\textwidth}
    \includegraphics[width=0.45\textwidth]{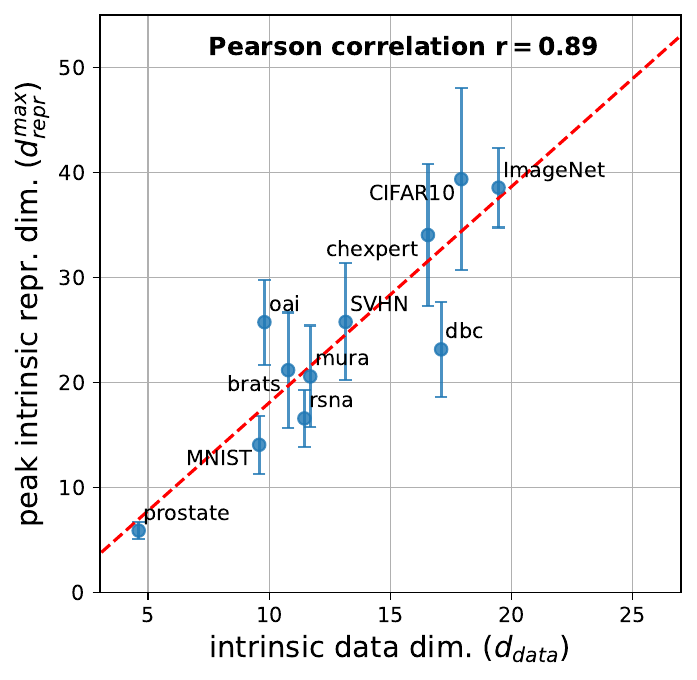}
    \caption{Correlation of peak hidden representation intrinsic dimension $\drmax$ with intrinsic dataset dimension $\dd$, averaged (with standard deviation error bars) over all models.}
    \label{fig:drpeak_vs_dd}
\end{wrapfigure}

When a network's representations reach a peak intrinsic dimension $\drmax$ at some depth (Fig. \ref{fig:humpbacks}), it implies that task-irrelevant features have fully been decorrelated from task-relevant features in the post-processing stage, such that the former can be removed from the representation in the remaining layers (the compression stage) without affecting the latter. We therefore hypothesize that $\drmax$ will be proportional to the dataset's $\dd$ because it describes the input data features in a decorrelated manner (as degrees of freedom) before any task-irrelevant features are discarded in later layers. We evaluated this empirically as shown in Fig. \ref{fig:drpeak_vs_dd}, averaging $\drmax$ and $\dd$ estimates for each dataset over all models, and intriguingly found this correlation to be quite high ($r=0.89$), with $\drmax$ being roughly double $\dd$. This relationship and correlation is quite consistent for different training set sizes (Appendix \ref{app:additionalresults:ddata}).

Why is it not the case that this constant of proportionality is closer to $1$, \ie, $\drmax\approx\dd$? We hypothesize that while the networks learn to compress input data to a dimensionality much lower than the \textit{extrinsic} dimension/pixel count (on the order of $10^5$) while still well-representing the data's information content, it is not quite the perfectly optimal, truly minimum number of degrees of freedom that can describe the data. While this could be due to limited network capacity, in our over-parameterized regime it is more likely that compressing the data further simply has a marginal effect on downstream task performance, as the representation dimensionality is already quite low and usable for downstream task prediction in the compression stage of the network, so the network has no need to learn to do so.

\section*{Related Work}

Beyond intrinsic dimension (ID), other metrics of the intrinsic complexity, dimensionality, capacity, or related characteristics of a hidden layer's representations include manifold capacity \cite{cohen2020separability}, linear ID \cite{huang2018mechanisms}, local ID \cite{amsaleg2015estimating,ma2018characterizing}, and others. We choose to analyze MLE estimator-based ID rather than manifold capacity due to it not requiring specifying labels/a task for the dataset, as manifold capacity focuses on how class representation manifolds are modified through the network, so that our findings may more easily generalize to models for tasks beyond classification. We focus on global nonlinear ID rather than local or linear ID due to its established relationship with important phenomena such as network generalization ability \cite{ansuini2019intrinsic}. Other general nonlinear manifold ID estimators beyond MLE include TwoNN \cite{facco2017estimating}, GeoMLE \cite{gomtsyan2019geometry}, and k-NN Graph Distance \cite{granata2016accurate}.

\section*{Conclusions}

In this work, we explored how the intrinsic dimension (ID) of hidden representations changes through a neural network's depth, in particular studying how the shape of this ID curve changes depending on the training dataset and data domain. We found that these curves peak at noticeably earlier layers when the model is trained with medical imaging data compared to natural images, implying a discrepancy between the two data domains in the abstractness of prediction-relevant features. We also studied the relationship between this peak representation ID and the ID of the dataset in its input space, between which we found a strong correlation, pointing to a close connection between the information content of a model's representations and the content of its training data. 

We expect that our findings will extend to other tasks beyond binary classification, not just because the ID estimator does not use labels, but also because similar ``pre-processing then compression'' representation ID progression curves (Fig. \ref{fig:humpbacks}) have been seen in models for both multi-class classification \cite{ansuini2019intrinsic} and other non-supervised, non-vision (as well as non-CNN) models \cite{valeriani2024geometry}. Future research directions could include (1) further probing our proposed connections between learned feature ``abstractness'' and peak representation ID, such as by seeing if training for classification tasks with increasingly abstract super-classes (\eg, as in WordNet \cite{miller1995wordnet}) results in an increased peak ID depth in the network; as well as (2) attempting to disentangle the effects of dataset ID and feature ``abstractness'' on peak representation ID using experiments that can control the precise ID of training data, such as in \cite{pope2021intrinsic}.

\bibliographystyle{plain}
\bibliography{main}

\appendix

\section{Supplementary Material}

\subsection{Additional Results}
\label{app:additionalresults}

\subsubsection{Progression of hidden representation intrinsic dimension through network depth.}
\label{app:additionalresults:humpbacks}

\begin{figure}[!htbp]
\centering
\includegraphics[width=0.99\textwidth]{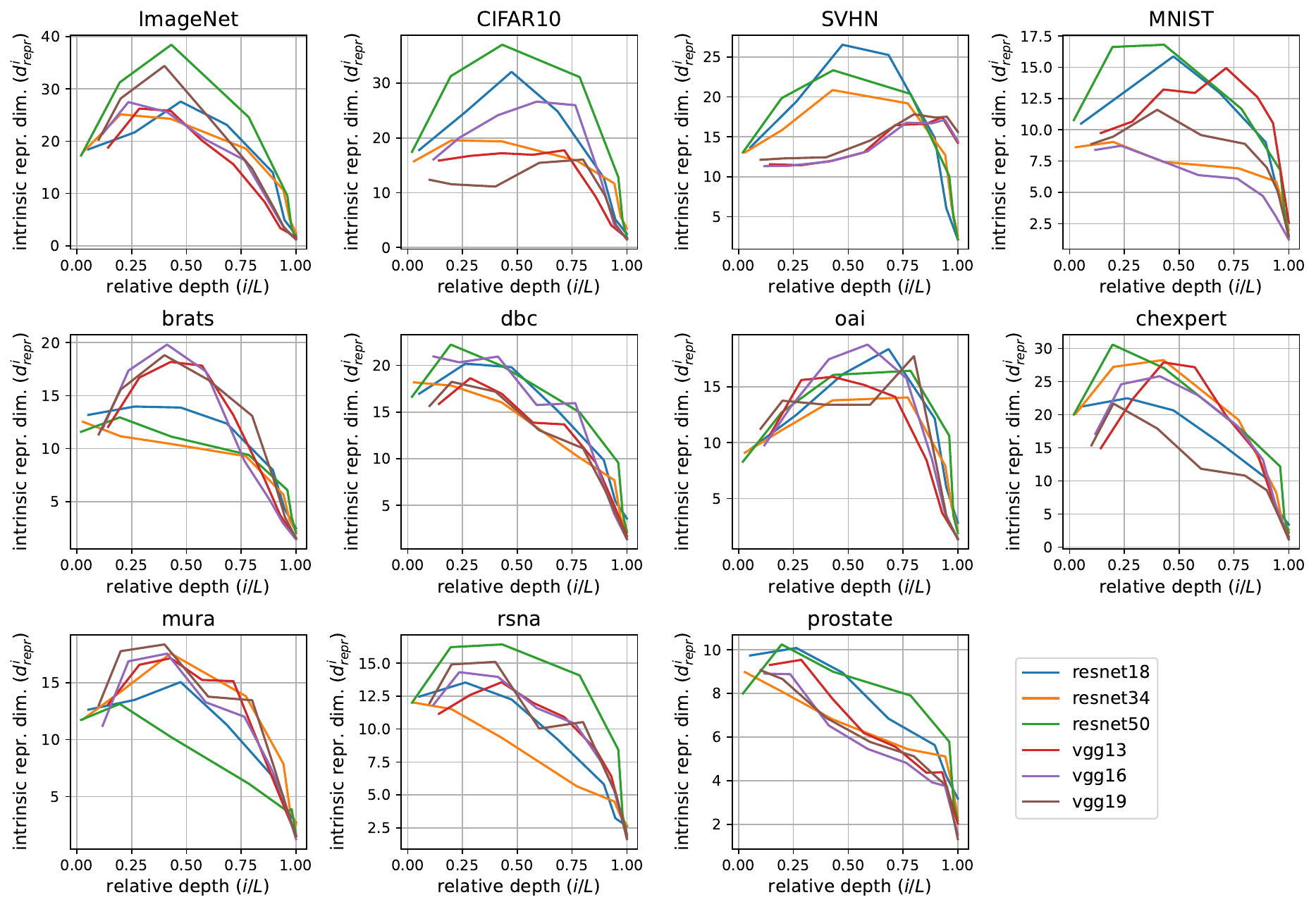}
\caption{Progression of hidden representation intrinsic dimension $\dr^i$ with respect to relative network depth $i/L$, for all models trained on each dataset, for $N=500$. \textbf{Top row:} natural image models. \textbf{Lower rows:} medical image models.}
\label{fig:humpbacks_500}
\end{figure}

\begin{figure}[!htbp]
\centering
\includegraphics[width=0.99\textwidth]{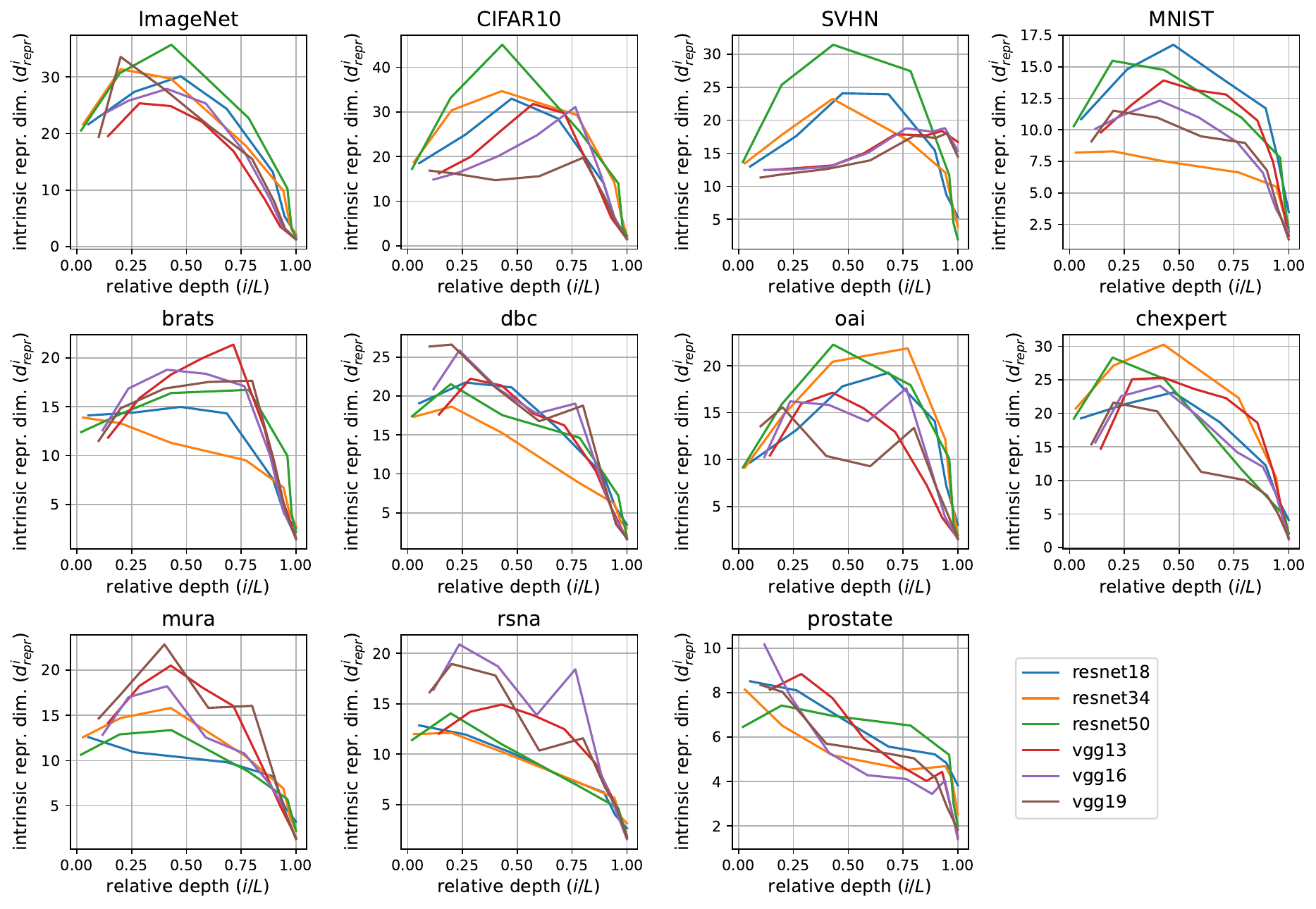}
\caption{Progression of hidden representation intrinsic dimension $\dr^i$ with respect to relative network depth $i/L$, for all models trained on each dataset, for $N=750$. \textbf{Top row:} natural image models. \textbf{Lower rows:} medical image models.}
\label{fig:humpbacks_750}
\end{figure}

\begin{figure}[!htbp]
\centering
\includegraphics[width=0.99\textwidth]{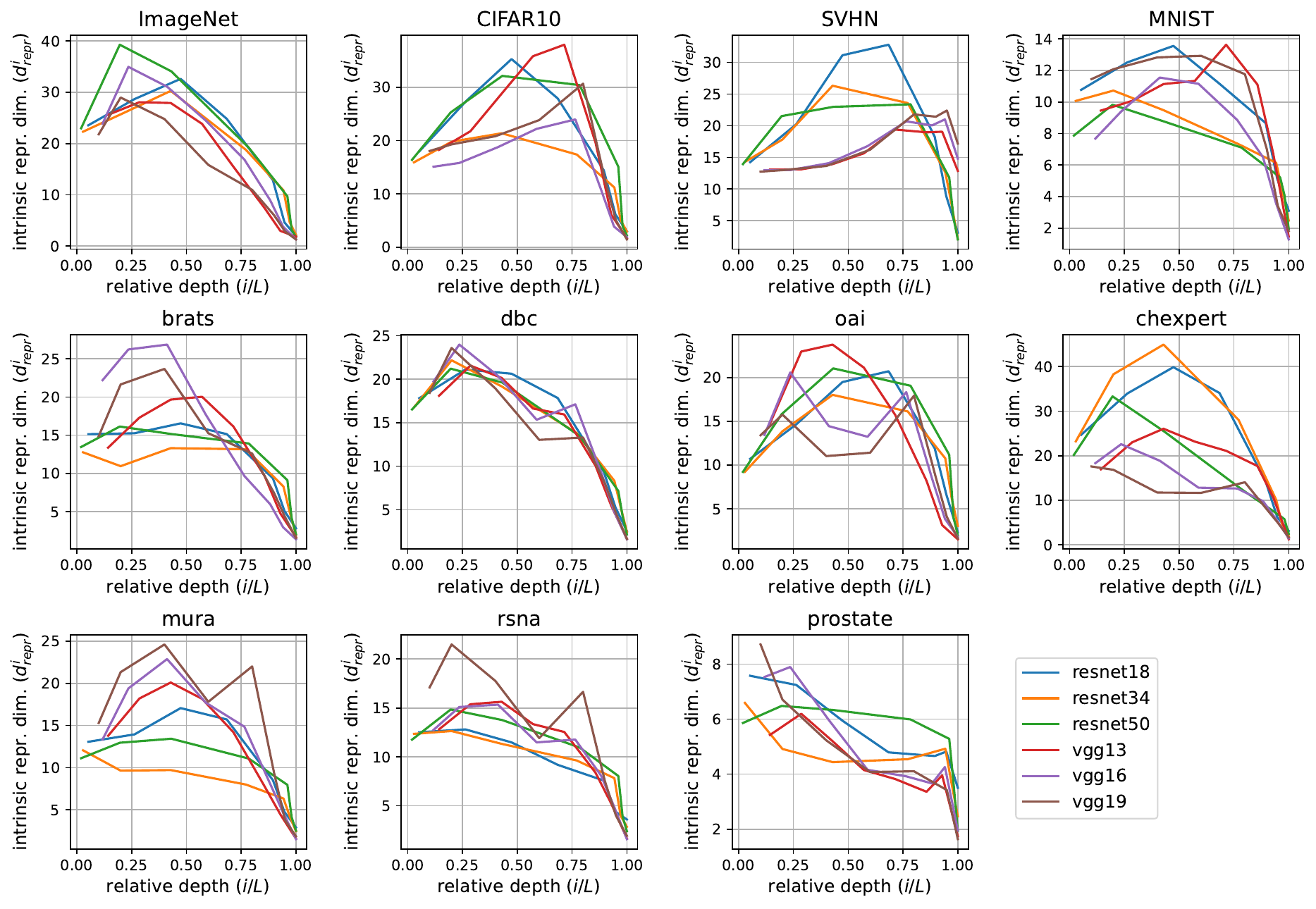}
\caption{Progression of hidden representation intrinsic dimension $\dr^i$ with respect to relative network depth $i/L$, for all models trained on each dataset, for $N=1000$. \textbf{Top row:} natural image models. \textbf{Lower rows:} medical image models.}
\label{fig:humpbacks_1000}
\end{figure}

\begin{figure}[!htbp]
\centering
\includegraphics[width=0.99\textwidth]{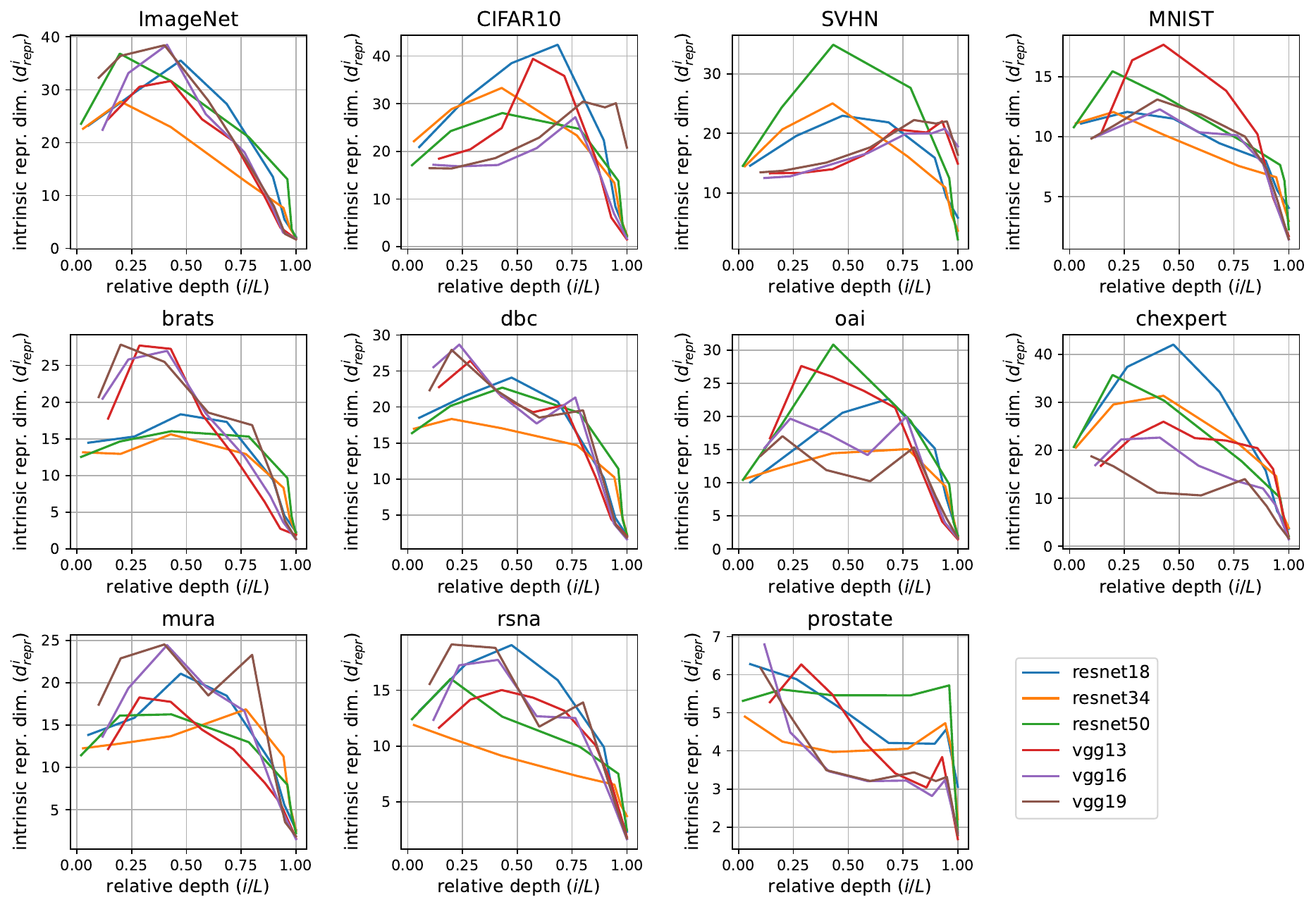}
\caption{Progression of hidden representation intrinsic dimension $\dr^i$ with respect to relative network depth $i/L$, for all models trained on each dataset, for $N=1250$. \textbf{Top row:} natural image models. \textbf{Lower rows:} medical image models.}
\label{fig:humpbacks_1250}
\end{figure}

\begin{figure}[!htbp]
\centering
\includegraphics[width=0.99\textwidth]{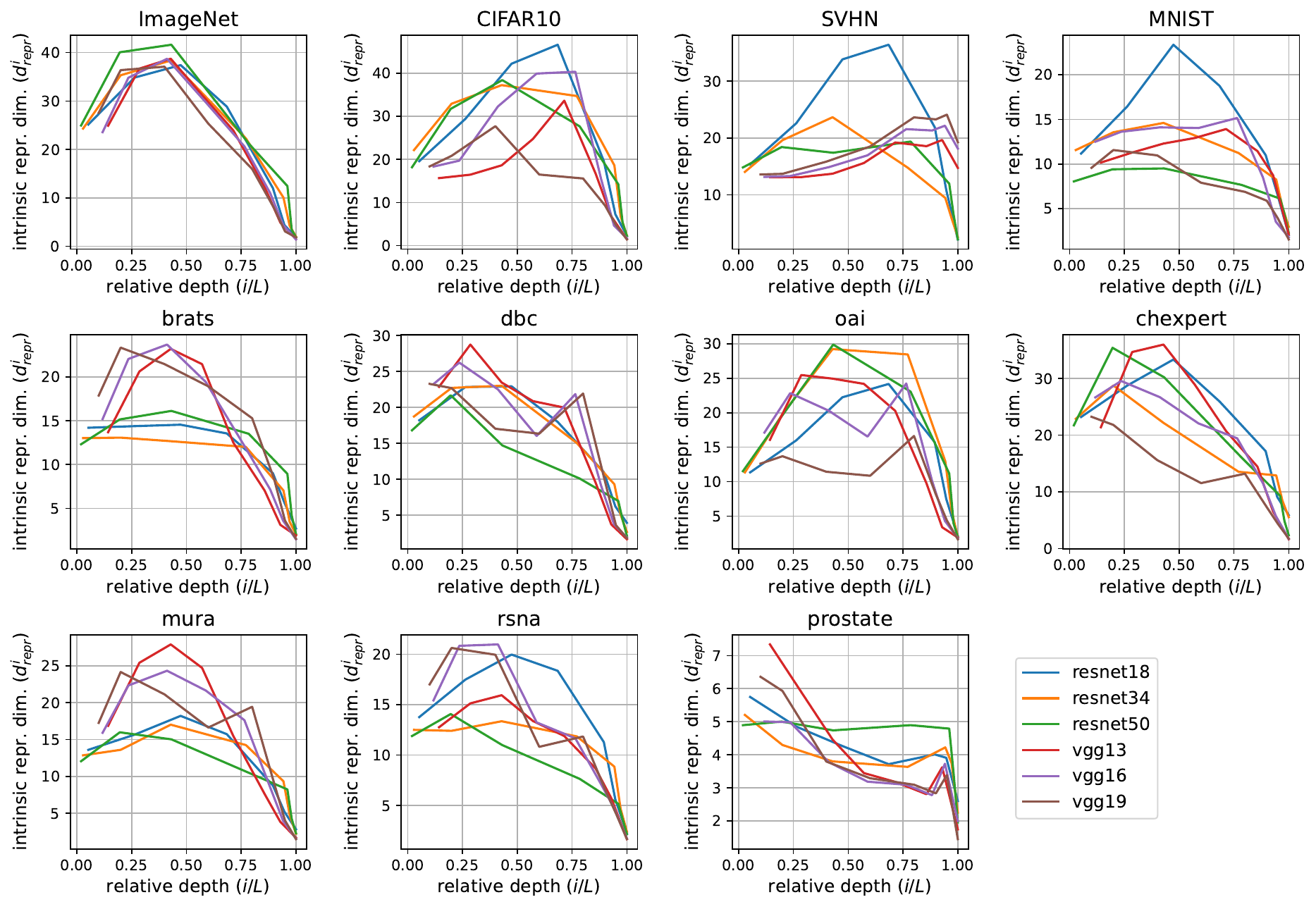}
\caption{Progression of hidden representation intrinsic dimension $\dr^i$ with respect to relative network depth $i/L$, for all models trained on each dataset, for $N=1500$. \textbf{Top row:} natural image models. \textbf{Lower rows:} medical image models.}
\label{fig:humpbacks_1500}
\end{figure}

We provide Table \ref{tab:peaks_app} to show the same results of average peak intrinsic dimension $\drmax$ and the relative network depth $i^*/L$ at which it was attained for natural vs. medical images, to supplement Table \ref{tab:peaks} for all training set sizes $N$. We also show the actual representation ID curve plots for each dataset for these other $N$ in Figs. \ref{fig:humpbacks_500}, \ref{fig:humpbacks_750}, \ref{fig:humpbacks_1000}, \ref{fig:humpbacks_1250}, and \ref{fig:humpbacks_1500} to supplement Fig. \ref{fig:humpbacks}.

\begin{table}[!htbp]
\centering
\small
\setlength{\tabcolsep}{4pt}
\begin{tabular}{l||cccccc}
    \multicolumn{7}{c}{\textbf{Natural Image Models}} \\
    & $N=500$ & $N=750$ & $N=1000$ & $N=1250$ & $N=1500$ & $N=1750$ \\
    \hline
    $\drmax$ & $22 \pm 6$ & $25 \pm 8$ & $25 \pm 8$ & $27 \pm 8$ & $29 \pm 10$  & $29 \pm 10$ \\
    $i^*/L$ &  $0.54 \pm 0.13$ & $0.53 \pm 0.17$ & $0.58 \pm 0.17$ & $0.54 \pm 0.16$  & $0.63 \pm 0.12$ & $0.57 \pm 0.15$ \\ \hline
\end{tabular}

\begin{tabular}{l||cccccc}
    \multicolumn{7}{c}{\textbf{Medical Image Models}} \\
    & $N=500$ & $N=750$ & $N=1000$ & $N=1250$ & $N=1500$ & $N=1750$ \\
    \hline
    $\drmax$ & $17 \pm 5$ & $18 \pm 5$ & $19 \pm 7$ & $20 \pm 7$  & $21 \pm 7$ & $21 \pm 8$ \\
    $i^*/L$ & $0.36 \pm 0.16$ & $0.37 \pm 0.16$ & $0.36 \pm 0.12$ & $0.38 \pm 0.10$ & $0.36 \pm 0.15$ & $0.35 \pm 0.17$ \\ \hline
\end{tabular}

\caption{Hidden representation peak intrinsic dimension $\drmax$ and the relative network depth $i^*/L$ at which it was attained, on average over all natural image models vs. all medical image models, for each training set size $N$ (extending the results of Table \ref{tab:peaks}, averaged over all architectures.}
\label{tab:peaks_app}
\end{table}

\subsubsection{The relationship between hidden representation intrinsic dimension and dataset intrinsic dimension.}
\label{app:additionalresults:ddata}

\begin{figure}[!htbp]
    \centering
    \includegraphics[width=0.3\textwidth]{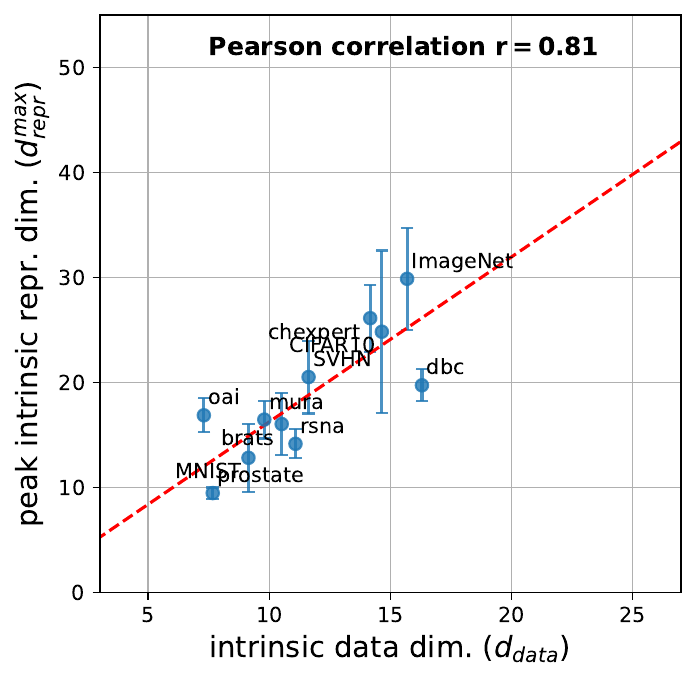}
    \includegraphics[width=0.3\textwidth]{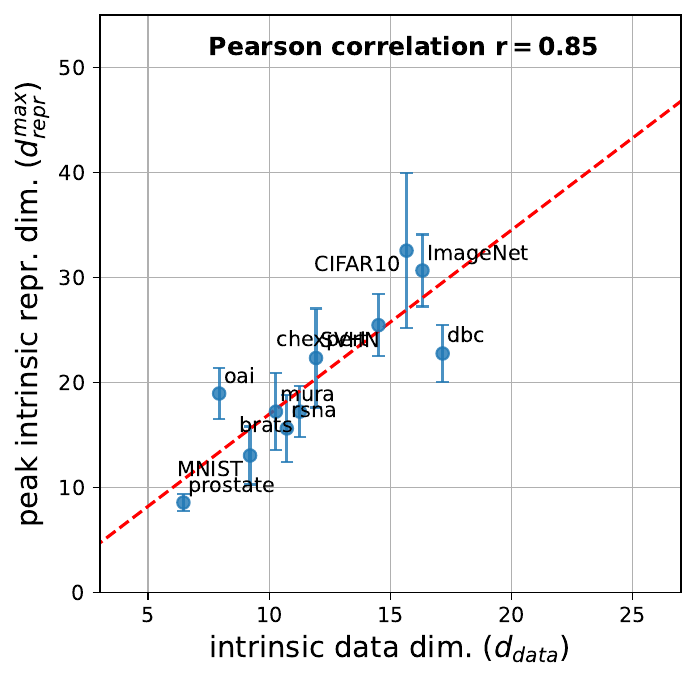}
    \includegraphics[width=0.3\textwidth]{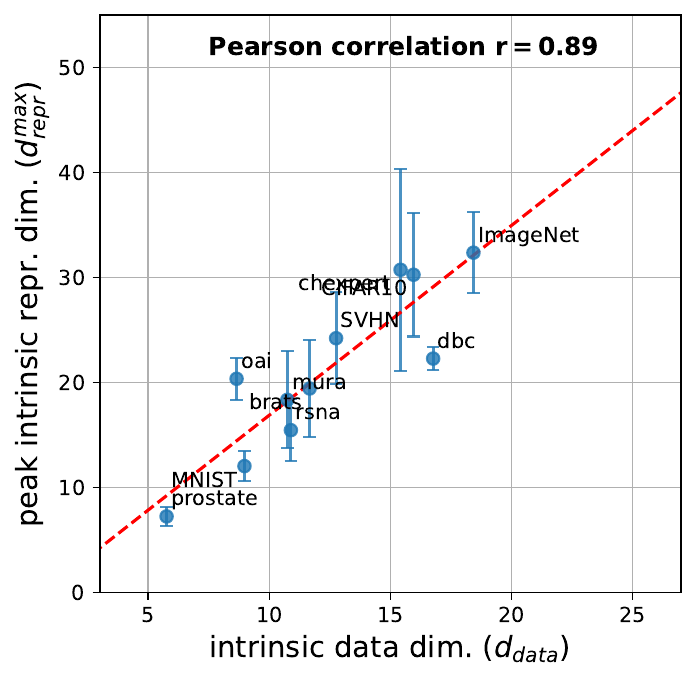}
    \includegraphics[width=0.3\textwidth]{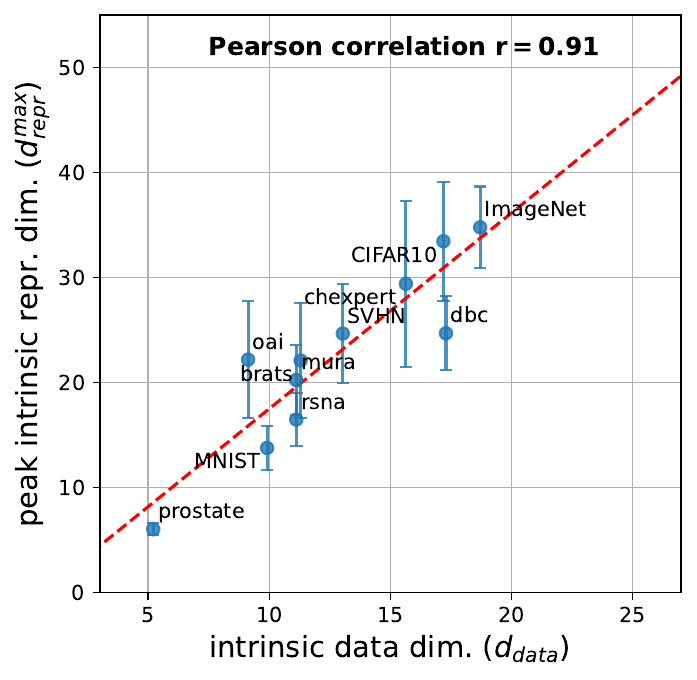}
    \includegraphics[width=0.3\textwidth]{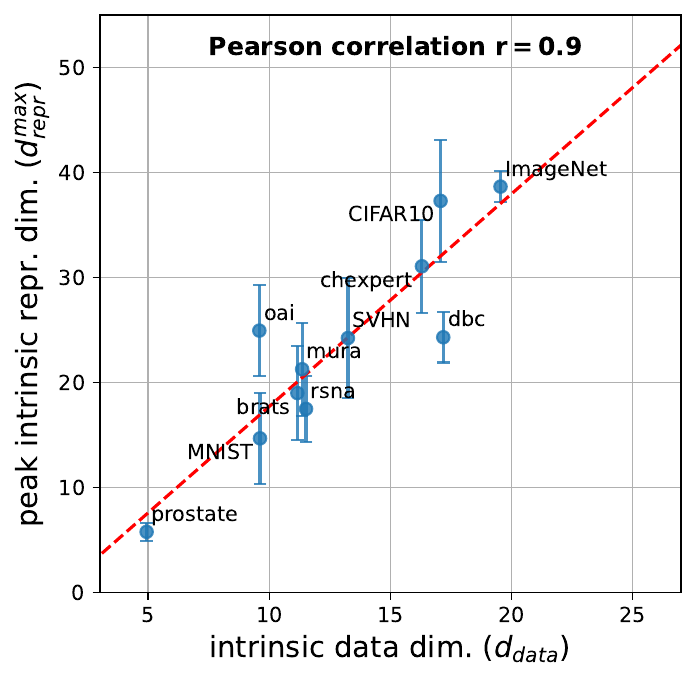}
    \caption{Correlation of peak hidden representation intrinsic dimension $\drmax$ with intrinsic dataset dimension $\dd$, averaged over all models, for additional training site sizes of $N=500, 750, 1000, 1250, 1500$, ordered left-to-right, to supplement the $N=1750$ result of Fig. \ref{fig:drpeak_vs_dd}.}
    \label{fig:drpeak_vs_dd_app}
\end{figure}

We provide Fig. \ref{fig:drpeak_vs_dd_app} to show the same results of Fig. \ref{fig:drpeak_vs_dd} for all other training set sizes.

\end{document}